\documentclass[10pt,twocolumn,letterpaper]{article}

\usepackage{cvpr}              

\usepackage{graphicx}
\usepackage{amsmath}
\usepackage{amssymb}
\usepackage{booktabs}
\usepackage{multirow}
\usepackage{tabularx}
\usepackage[ruled]{algorithm2e}
\usepackage[accsupp]{axessibility}  
\usepackage[pagebackref,breaklinks,colorlinks]{hyperref}

\usepackage[capitalize]{cleveref}
\crefname{section}{Sec.}{Secs.}
\Crefname{section}{Section}{Sections}
\Crefname{table}{Table}{Tables}
\crefname{table}{Tab.}{Tabs.}

\begin{document}


\title{Stacked Hybrid-Attention and Group Collaborative Learning\\
for Unbiased Scene Graph Generation}

\author{Xingning Dong\textsuperscript{\rm 1,\rm 2}, \quad Tian Gan\textsuperscript{\rm 1}{\footnotemark[2]}, \quad Xuemeng Song\textsuperscript{\rm 1}, \quad Jianlong Wu\textsuperscript{\rm 1}, \quad Yuan Cheng\textsuperscript{\rm 2}{\footnotemark[2]}, \quad Liqiang Nie\textsuperscript{\rm 1}\\[2pt]
\normalsize{\textsuperscript{\rm 1}Shandong University, \quad \quad \textsuperscript{\rm 2}Ant Group}\\
{\tt\small dongxingning1998@gmail.com, \quad gantian@sdu.edu.cn, \quad sxmustc@gmail.com} 
\\
{\tt\small jlwu1992@sdu.edu.cn, \quad chengyuan.c@antgroup.com, \quad nieliqiang@gmail.com}
}


\maketitle
\renewcommand{\thefootnote}{\fnsymbol{footnote}}
\footnotetext[2]{Corresponding authors.}

\begin{abstract}
Scene Graph Generation, which generally follows a regular encoder-decoder pipeline, aims to first encode the visual contents within the given image and then parse them into a compact summary graph. Existing SGG approaches generally not only neglect the insufficient modality fusion between vision and language, but also fail to provide informative predicates due to the biased relationship predictions, leading SGG far from practical. Towards this end, 
we first present a novel Stacked Hybrid-Attention network, which facilitates the intra-modal refinement as well as the inter-modal interaction, to serve as the encoder. We then devise an innovative Group Collaborative Learning strategy to optimize the decoder. Particularly, based on the observation that the recognition capability of one classifier is limited towards an extremely unbalanced dataset, we first deploy a group of classifiers that are expert in distinguishing different subsets of classes, and then cooperatively optimize them from two aspects to promote the unbiased SGG. Experiments conducted on VG and GQA datasets demonstrate that, we not only establish a new state-of-the-art in the unbiased metric, but also nearly double the performance compared with two baselines. Our code is available at \href{https://github.com/dongxingning/SHA-GCL-for-SGG}{https://github.com/dongxingning/SHA-GCL-for-SGG}.

\end{abstract}


\begin{figure}[t]
	\centering
	\includegraphics[width=0.47\textwidth]{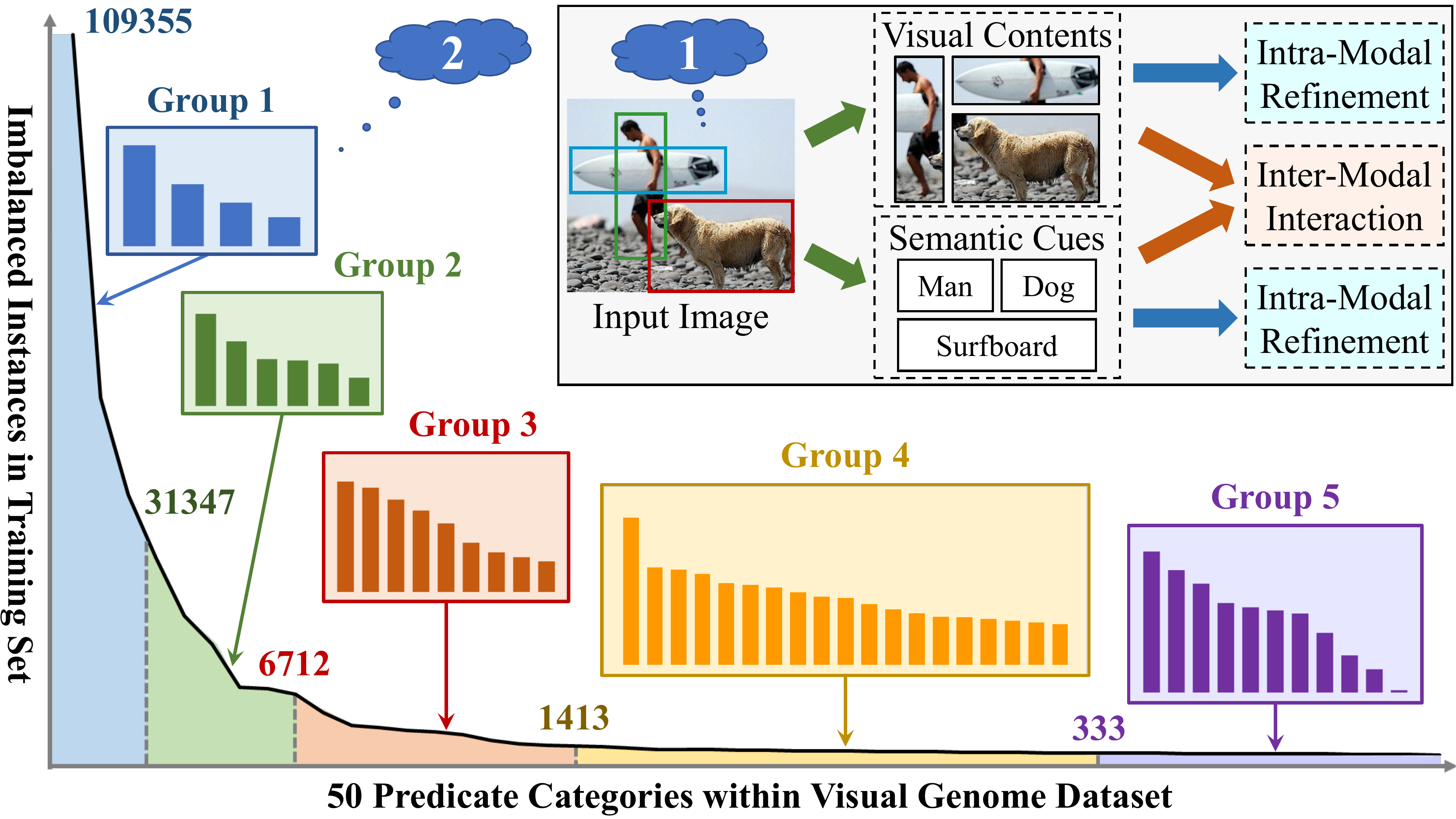}
	\vspace{-0.2cm}
	\caption{Two intentions to promote the unbiased SGG. (1) For the insufficient modality fusion, we aim to enhance both the intra-modal refinement and the inter-modal interaction (see the top-right corner of the figure). And (2) we split the extremely unbalanced dataset into a set of relatively balanced groups, based on which we configure the classification space for all the newly-added classifiers (see the rest part of the figure).}
	\vspace{-0.3cm}
	\label{introduction}
\end{figure}

\section{Introduction}
\label{sec:intro}

Scene Graph Generation (SGG) \cite{xu2020survey} targets at organizing all the objects and their pairwise relationships into a compact summary graph. As an intermediate visual understanding task, SGG could benefit various vision-and-language tasks, including cross-modal retrieval\cite{guo2020visual,song2021spatial,dhamo2020semantic}, image captioning\cite{zhong2020comprehensive,chen2020say,gu2019unpaired}, and visual question answering\cite{teney2017graph,zhang2019empirical,hildebrandt2020scene}. However, SGG is still far from satisfactory for practical applications due to the insufficient modality fusion and the biased relationship predictions. 

Though it is manifestly proved that incorporating semantic cues (language priors of object class names) into visual contents (object proposals) could significantly improve the generation capability\cite{lu2016visual, liang2018visual}, most of the recent approaches\cite{zellers2018neural,tang2019learning,tang2020unbiased,yan2020pcpl,yu2020cogtree,li2021bipartite,zareian2020bridging,sharifzadeh2020classification} simply fuse these visual and semantic features by summing up directly or concatenation, which limits the model to further infer their interaction information. To address this under-explored insufficient modality fusion between visual contents and semantic cues, we aim to strengthen the encoder via jointly exploring the intra-modal refinement and the inter-modal interaction, as illustrated in Figure \ref{introduction}. To implement this intention, we first design the Self-Attention (SA) unit and the Cross-Attention (CA) unit to capture the intra-modal and inter-modal information, respectively. We then organize these two units into a Hybrid-Attention (HA) layer, and stack several HA layers to build the encoder. The proposed Stacked Hybrid-Attention (SHA) network could adequately explore the multi-modal interaction, thus improving the relationship prediction performance.

The other prominent issue faced by existing SGG methods is the biased relationship predictions due to the long-tailed data distribution. Since only a few head predicates (\textit{e}.\textit{g}., \textit{on}, \textit{has}) possess massive and various instances, they would dominate the training procedure and lead the output scene graphs with few informative tail predicates (\textit{e}.\textit{g}., \textit{riding}, \textit{watching}), which could hardly support a wide range of downstream tasks. Though various debiasing approaches\cite{suhail2021energy, chiou2021recovering, wen2020unbiased} have been proposed, they are vulnerable to over-fitting the tail classes and sacrificing much on the head ones, leading to the other extreme. In a sense, we conjecture that this dilemma may root in the fact that a naive SGG model, regardless of the conventional or debiasing one, could only differentiate a limited range of predicates whose amount of training instances are relatively equal.

Intuitively, since a single classifier struggles in achieving a reasonable prediction trade-off, we can divide the biased predicate classes into several balanced subsets, then introduce more classifiers to conquer each of them, and ultimately leverage these classifiers to cooperatively address this challenge. To fulfill this ``divide-conquer-cooperate'' intuition, we propose the Group Collaborative Learning (GCL) strategy, where we 1) \textbf{first divide}: As a single classifier is adequate to differentiate the classes within a balanced dataset, we first divide all the predicates into a set of relatively balanced groups according to their amount of training instances, as illustrated in Figure \ref{introduction}. 2) \textbf{Then conquer}: We then borrow the idea from the class-incremental learning \cite{hu2020learning} to force all the classifiers to follow a continuously growing classification space, \ie, each classifier would extend the previous classification space by incorporating a newly-added group of predicates. Besides, we devise the Median Re-Sampling strategy to provide each classifier with a relatively balanced training set. Based on this group-incremental configuration, these nested classifiers could fairly treat the predicates within their classification space, thus they would be more likely to learn the discriminating representations, especially towards the newly-added group. 3) \textbf{Ultimately cooperate}: We further leverage these classifiers to cooperatively enhance the unbiased relationship predictions from two aspects. First, we propose the Parallel Classifier Optimization (PCO) to jointly optimize all the classifiers. This can be seen as a ``weak constraint'', since we expect that gathering all the gradients could promote the recognition capability of each classifier. Second, we devise the Collaborative Knowledge Distillation (CKD) to ensure that the discriminating capability learned previously could be well translated to the subsequent classifiers. This can be seen as a ``strong constraint'', since we force each classifier to mimic the prediction behavior from its predecessors. By employing these two constraints, we effectively mitigate the overwhelming punishments to the tail classes as well as compensate for the under-fitting on the head ones.

The contributions of our work are three-folds:

\vspace{-0.2cm}
\begin{itemize}
\setlength{\itemsep}{0pt}
\setlength{\parsep}{0pt}
\setlength{\parskip}{0pt}

	\item We present a novel Stacked Hybrid Attention network to strengthen the encoder in SGG, which addresses the under-explored insufficient modality fusion problem.
	
	\item We design the Group Collaborative Learning strategy to optimize the decoder in SGG. Particularly, we deploy a group of classifiers and cooperatively optimize them from two aspects, thus effectively addressing the intractable biased relationship prediction problem.
	
	\item Experiments conducted on VG and GQA dataset indicate that, we not only establish a new state-of-the-art in the unbiased metric, but also nearly 
	double the performance compared with two typical baselines when employing our model-agnostic GCL.

\end{itemize}
\vspace{-0.2cm}

\begin{figure*}[t]
	\begin{center}
		\includegraphics[width=1\textwidth]{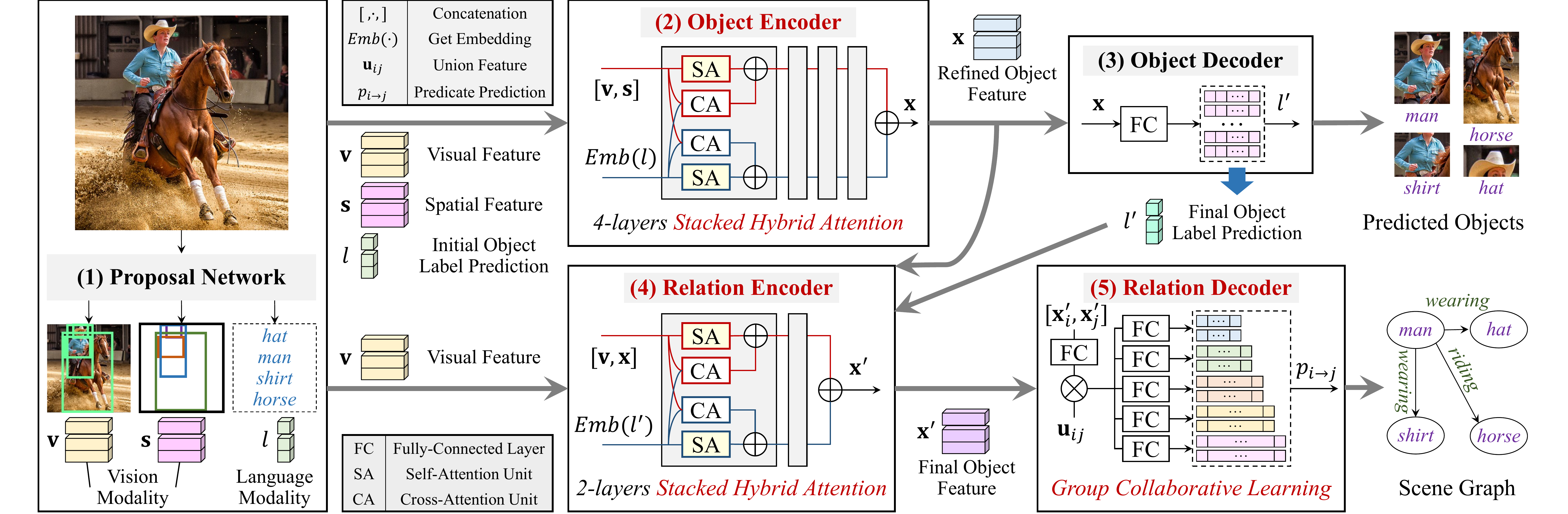}
	\end{center}
	\vspace{-0.4cm}
	\caption{The framework of the common pipeline in SGG, which includes five key components. Notably, we improve three key components marked in red in the figure. Specifically, we propose the Stacked Hybrid-Attention network to enhance the object encoder and the relation encoder, and we also devise the Group Collaborative Learning strategy to guide the training of the relation decoder.}
	
	\vspace{-0.4cm}
	\label{framework}
\end{figure*}

\section{Related Work}
\noindent\textbf{Scene Graph Generation.}
SGG provides an efficient way for scene understanding by decoding the visual relationships into a summary graph. Early approaches\cite{lu2016visual,dai2017detecting,liang2018visual,liao2019natural} were mainly dedicated to incorporating more features from various modalities, but they neglected the rich visual context, leading to sub-optimal performance. In order to tackle such deficiency, later approaches employed more powerful feature refinement modules to encode the rich contextual information, such as message passing strategy\cite{xu2017scene,li2021bipartite}, sequential LSTMs\cite{zellers2018neural,tang2019learning}, graph neural networks\cite{chen2019knowledge,zareian2020bridging}, and self-attention networks\cite{lin2020gps,sharifzadeh2020classification}. Though the performance is improved in the regular metrics, the relations they predicted are often trivial and less informative due to the biased training data, which could hardly support the downstream vision-and-language tasks. Therefore, various approaches\cite{suhail2021energy, chiou2021recovering, wen2020unbiased} have been proposed to tackle the biased relationship predictions, including employing debiasing strategies like re-sampling\cite{li2021bipartite} or re-weighting\cite{yan2020pcpl}, disentangling unbiased representations from the biased\cite{tang2020unbiased}, and utilizing the tree structure to filter the irrelevant predicates\cite{yu2020cogtree}. 
However, these approaches are vulnerable to over-fitting on the tail classes with much sacrifice on the head ones. Based on the observation that a single classifier could hardly differentiate all the classes within a biased dataset, and inspired by the ``divide-conquer-cooperate'' intuition, we propose the Group Collaborative Learning strategy to guide the training of the decoder. In this way, we not only significantly improve the prediction performance towards the tail classes, but also effectively preserve the discriminating capability learned by the head ones, thus achieving a reasonable prediction trade-off. 

\noindent\textbf{Cross-attention Models.} Research towards improving multi-modal fusion\cite{wei2019neural, wei2019mmgcn} and building cross-attention models\cite{zheng2021deep,wiles2021co} have been attracting increasing interest in various vision-and-language tasks. For example, Yu \textit{et} \textit{al.}\cite{yu2019deep} proposed the deep Modular Co-Attention Network to fully model the interaction between question words and image regions in VQA, and Lu \textit{et} \textit{al.}\cite{lu2019vilbert} proposed ViLBERT to extend BERT architecture for jointly pre-training images and texts. Nevertheless, few of the approaches in SGG dedicate to addressing the insufficient modality fusion between object proposals and their corresponding class names. Therefore, we propose the Stacked Hybrid-Attention (SHA) network to facilitate both the intra-modal refinement and the inter-modal interaction.

\noindent\textbf{Knowledge Distillation.} Knowledge distillation\cite{gou2021knowledge,hinton2015distilling,mirzadeh2020improved} aims to distill the knowledge from a larger deep network into
a small one, which is widely employed in various tasks, including model compression\cite{wang2019private, bai2020few}, label smoothing\cite{yuan2020revisiting, shen2021label}, and data augmentation\cite{gordon2019explaining, feng2021learning}. Note that the conventional knowledge distillation approaches generally follow a teacher-student pipeline. These two networks are optimized in different time steps as the teacher network is usually available beforehand. Different from this model-to-model paradigm, after adding several classifiers, we allow the previous classifiers to generate the outputs as soft labels to constrain the training of the subsequent, thus establishing a layer-to-layer ``knowledge transfer''.


\section{Methodology}

\subsection{Problem Formulation} 
\label{Sec31}
SGG aims to generate a summary graph $\mathcal{G}$ that highly generalizes the contents of a given image $\boldsymbol{I}$. Towards this end, we first detect all the objects within the image $\boldsymbol{I}$, denoted as $\mathcal{O}=\{o_{i}\}_{i=1}^N$. Then for each object pair ($o_{i}$, $o_{j}$), we predict its predicate $p_{i\rightarrow j}$. Ultimately, we organize all these predictions in the form of triplets to construct the scene graph, which can be formulated as $\mathcal{G} = \{(o_i,p_{i\rightarrow j},o_j)|o_i, o_j\in \mathcal{O}, p_{i\rightarrow j}\in \mathcal{P}\}$, where $\mathcal{P}$ stands for the set of all the possible predicates.

\subsection{Overall Framework}
\label{Sec32}
As illustrated in Figure \ref{framework}, our framework is based on the common pipeline followed by typical SGG  approaches\cite{zellers2018neural,tang2019learning,yan2020pcpl,yu2020cogtree}, which is a regular encoder-decoder structure.

\textbf{Proposal Network} is actually a pre-trained object detector. Given an image $\boldsymbol{I}$, it generates a set of object predictions $ \mathcal{O}=\{o_{i}\}_{i=1}^N$. For each object $o_{i}$, it provides a visual feature $\mathbf{v}_i$, a spatial feature $\mathbf{s}_i$ of the bounding box coordinates, and an initial object label prediction ${l}_i$.

\textbf{Object Encoder} aims to obtain the refined object feature $\mathbf{x}_i$ for further predictions, which is calculated as:
\begin{equation}
	\mathbf{x}_i = {Enc}^{obj}([\mathbf{v}_i, FC(\mathbf{s}_i)], Emb( {l}_i) ),
\end{equation}
where ${Enc}^{obj}(\cdot) $ represents the object encoder, which can be any feature refinement modules (\textit{e}.\textit{g}., BiLSTMs~\cite{zellers2018neural} and GNNs\cite{chen2019knowledge}), $[,\cdot,]$ denotes the concatenation operation, $FC(\cdot) $ represents a fully-connected layer, and $Emb(\cdot) $ refers to a pre-trained language model to acquire the semantic feature of $ o_{i}$ based on its initial object label prediction ${l}_i$.

\textbf{Object Decoder} aims to obtain the final object label prediction ${l}^{\prime}_i$ based on the refined object feature $\mathbf{x}_i$, which is calculated as:
\begin{equation}
	{l}^{\prime}_i = {\rm argmax}({\rm Softmax}(Dec^{obj}(\mathbf{x}_i))),
\end{equation}
where ${Dec}^{obj}(\cdot) $ represents the object decoder, which is a single fully-connected layer.

\textbf{Relation Encoder} works on obtaining the final object feature $\mathbf{x}^{\prime}_i$ for predicate predictions, which is calculated as:
\begin{equation}
	\mathbf{x}^{\prime}_i = {Enc}^{rel}([\mathbf{v}_i, \mathbf{x}_i], Emb( {l}^{\prime}_i)),
\end{equation}
where ${Enc}^{rel}(\cdot) $ represents the relation encoder, which shares the same architecture with the object encoder. 

\textbf{Relation Decoder} is responsible for predicting the predicate label ${p}_{i\rightarrow j}$ based on the final object features of subject ${o}_i$ and object $o_j$, which is calculated as:
\begin{equation}
	{p}_{i\rightarrow j} = {\rm argmax}({\rm Softmax}({Dec}^{rel}(\mathbf{x}^{\prime}_i , \mathbf{x}^{\prime}_j , \mathbf{u}_{ij})) ,
\end{equation}
where ${Dec}^{rel}(\cdot) $ represents the relation decoder. We also follow \cite{zellers2018neural} to employ the union feature $\mathbf{u}_{ij}$ of the object pair ($o_i$, $o_j$) to enhance the predicate predictions.

It is worth noting that we improve three key components marked in red in Figure \ref{framework} to promote the unbiased SGG. Specifically, for the object encoder and the relation encoder, we propose the Stacked Hybrid-Attention (SHA) network to alleviate the insufficient modality fusion problem. Regarding the relation decoder, we devise the Group Collaborative Learning (GCL) strategy to address the intractable biased relationship prediction problem.

\begin{figure}[t]
	\centering
	\includegraphics[width=0.49\textwidth]{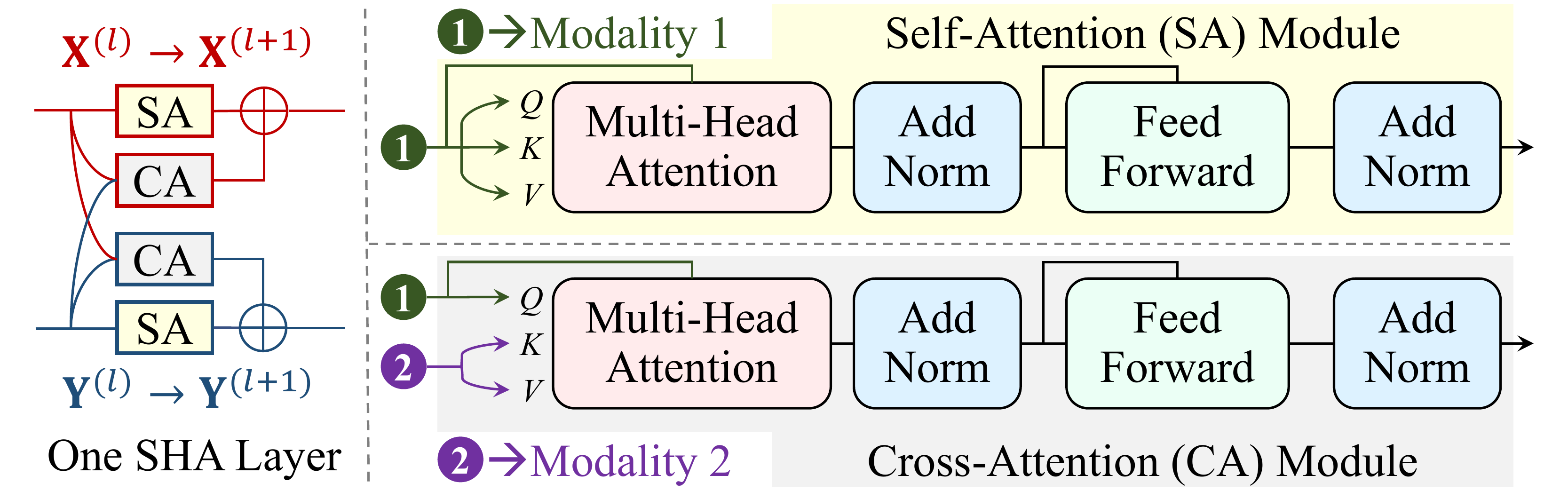}
	\vspace{-0.4cm}
	\caption{One single Stacked Hybrid-Attention (SHA) layer is composed of two types of attention units, \ie, Self-Attention (SA) unit to facilitate the intra-modal refinement and Cross-Attention (CA) unit to promote the inter-modal interaction.}
	\vspace{-0.4cm}
	\label{encoder}
\end{figure}

\begin{figure*}[t]
	\begin{center}
		\includegraphics[width=1\textwidth]{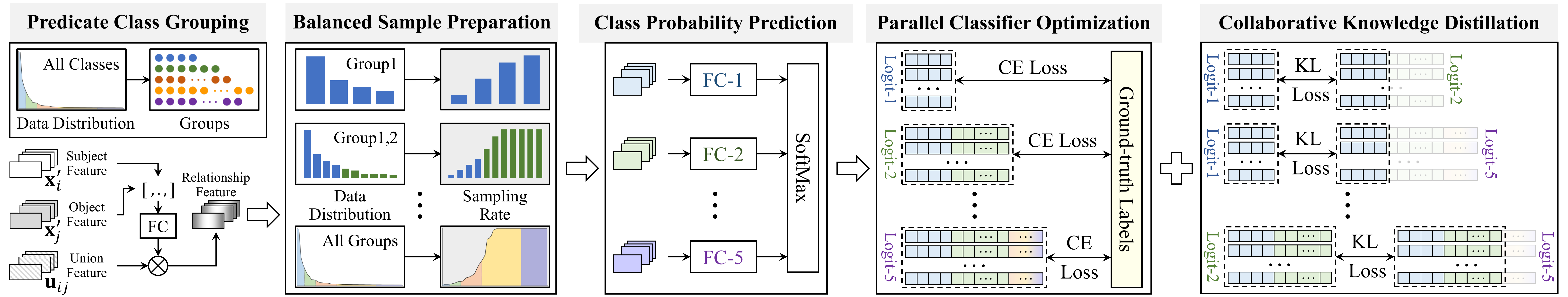}
	\end{center}
	\vspace{-0.4cm}
	\caption{Illustration of the proposed Group Collaborative Learning (GCL) strategy, which includes five key steps. It is worth noting that we design two optimization mechanisms, namely Parallel Classifier Optimization (PCO) and Collaborative Knowledge Distillation (CKD), to jointly guide the training of the relation decoder.}
	\vspace{-0.4cm}
	\label{decoder}
\end{figure*}

\subsection{Encoder: Stacked Hybrid-Attention}
\label{Sec33}
Beyond understanding the visual contents (object proposals) of a given image, the semantic cues (refer to the class names in SGG) are also indispensable for robust relationship predictions. Unfortunately, most of the approaches in SGG simply fuse these two modal features by summing up directly or concatenation, which may be insufficient to mine the underlying inter-modal interaction, thus resulting in sub-optimal performance. To address this deficiency, we propose the Stacked Hybrid Attention (SHA) network, which is composed of several SHA layers. Each SHA layer contains two parallel Hybrid-Attention (HA) cells, and each HA cell is a composition of two types of attention units, \ie, the Self-Attention (SA) unit to facilitate the intra-modal refinement, and the Cross-Attention (CA) unit to model the inter-modal interaction. As shown in Figure \ref{encoder}, both the SA unit and CA unit are built upon a multi-head attention module and a feed-forward module based on the attention mechanism\cite{vaswani2017attention}. The difference between SA and CA is whether the input features belong to the same modality.

Ultimately, we build our SHA network by cascading $L$ SHA layers in sequential order. For the $l$-th SHA layer, the feature propagation process can be formulated as:
\begin{equation}
    \left \{
    \begin{aligned}
	\mathbf{X}^{(l)} = SA(\mathbf{X}^{(l-1)}) + CA(\mathbf{X}^{(l-1)}, \mathbf{Y}^{(l-1)}),
	\\
	\mathbf{Y}^{(l)} = SA(\mathbf{Y}^{(l-1)}) + CA(\mathbf{Y}^{(l-1)}, \mathbf{X}^{(l-1)}),
	\end{aligned}
	\right.
\end{equation}
where $SA(\cdot)$ and $CA(\cdot)$ denote the self-attention and cross-attention computation, respectively. For the first SHA layer, we set its input feature $\mathbf{X}^{(0)} = \mathbf{X}$ and $\mathbf{Y}^{(0)} = \mathbf{Y}$, where $\mathbf{X}$ and $\mathbf{Y}$ denote the original visual feature and semantic feature, respectively. After obtaining the final visual feature $\mathbf{X}^{(L)}$ and semantic feature $\mathbf{Y}^{(L)}$ generated by the last SHA layer, we sum them up to get the refined output, which contains rich multi-modal interaction information.

\subsection{Decoder: Group Collaborative Learning}
\label{Sec34}

As aforementioned, when facing an extremely unbalanced dataset, a naive SGG model could hardly achieve a satisfactory prediction performance on all the predicate classes. Towards this end, we aim to deploy several classifiers which are expert in distinguishing different subsets of predicates, and organize these classifiers to cooperatively address the biased relationship predictions. Based on this ``divide-conquer-cooperate'' intention, we propose the Group Collaborative Learning (GCL) strategy. As shown in Figure \ref{decoder}, GCL contains five key steps as follows:

\textbf{Predicate Class Grouping} aims to split the unbalanced dataset into several relatively balanced groups, and then configure the classification space for all the classifiers. Based on the observation that the recognition capability would suffer from the biased data distribution, we aim to provide each classifier with a relatively balanced training set, thus it could adequately learn the discriminating representations towards a subset of predicates. Therefore, We first sort the predicate classes by their amount of training instances in descending order, obtaining a sorted set $\mathcal{P}_{all}=\{p_{i}\}_{i=1}^M$. We then divide $ \mathcal{P}_{all}$ into $K$ mutually exclusive groups $\{\mathcal{P}_{k}\}_{k=1}^K $ according to the pre-defined threshold $\mu$. The workflow is summarized in Algorithm \ref{BCD_alg}, where $Count({p}_{i})$ denotes the amount of training instances towards the predicate $p_{i}$. Line 3 in Algorithm \ref{BCD_alg} ensures that, for each group $\mathcal{P}_{k}$, the maximal amount of training instances will be no more than $\mu$ times of the minimal amount, thus the predicates in $\mathcal{P}_{k}$ share a relatively equal amount.

\begin{algorithm}
	\caption{Predicate Class Grouping.}
	\label{BCD_alg}
	\LinesNumbered
	\KwIn{A sorted predicate set $\mathcal{P}_{all}=\{p_{i}\}_{i=1}^M$, $\mu$}
	\KwOut{$K$ mutually exclusive groups $\{\mathcal{P}_{k}\}_{k=1}^K  $}
	Set $cur=1$, $k=1$, and $\mathcal{P}_{1} = \{\}$\;
	\For{$i\leftarrow 1$ \KwTo $M$}{
	    \If{$Count({p}_{cur}) > \mu * Count({p}_{i})$}{
	        $cur = i$\; 
	        $k = k+1$\;
			Set $\mathcal{P}_{k} = \{\}$\;
		}
		$\mathcal{P}_{k}= \mathcal{P}_{k} \cup \{{p}_{i} \}$
	}
\end{algorithm}

We then borrow the idea from the class-incremental learning \cite{hu2020learning}, and deploy a set of classifiers $ \{\mathcal{C}_{k}\}_{k=1}^{K}$ which follow a continuously growing classification space. Except for the first classifier $ \mathcal{C}_{1}$, other classifiers should recognize the predicate classes from both previous and current groups, \ie, the classification space in $ \mathcal{C}_{k}$ is $ \mathcal{P}_{k}^{\prime}= \mathcal{P}_{1} \cup \mathcal{P}_{2} \cup \cdots \cup \mathcal{P}_{k}$. Note that we only choose the last classifier $ \mathcal{C}_{K}$ to obtain the final predicate predictions in the evaluation stage.

\textbf{Balanced Sample Preparation} aims to achieve several balanced training sets provided for further joint optimization by re-sampling the instances. For each classifier $\mathcal{C}_{k}$ that incorporates a newly-added group $\mathcal{P}_{k}$ to extend the previous classification space $\mathcal{P}_{k-1}^{\prime}$ as $ \mathcal{P}_{k}^{\prime}=\mathcal{P}_{k} \cup \mathcal{P}_{k-1}^{\prime}$, we expect it could adequately learn the discriminating representations towards the predicates, particularly within the newly-added group $\mathcal{P}_{k}$. Therefore, for the predicates in the group $\mathcal{P}_{k}$, we should retain all of its training instances to facilitate the convergence. And for the predicates in the previous classification space $ \mathcal{P}_{k-1}^{\prime}$, since they have more samples in the original dataset, we should under-sample their training instances to avoid biased predictions.

To implement this intention, we propose the Median Re-Sampling strategy to perform the re-sampling operation. For each classification space $ \mathcal{P}_{k}^{\prime}$, we first calculate the median amount $Med({\mathcal{P}}_{k}^{\prime})$ over all the classes within  $ \mathcal{P}_{k}^{\prime}$. For example, if $\mathcal{P}_{k}^{\prime}$ is sorted in descending order and contains 9 predicate classes, the median amount $Med({\mathcal{P}}_{k}^{\prime})$ is equal to $Count({p}_{5})$. Then for each predicate class $p_{i}^{k}$ in  $ \mathcal{P}_{k}^{\prime}$, we calculate the sampling rate $\phi_{i}^{k}$ as follows:

\vspace{-0.4cm}
\begin{equation}
    \phi_{i}^{k} = 
	\left\{
	\begin{aligned}
		& \frac{Med({\mathcal{P}}_{k}^{\prime})}{ Count({p}_{i}) },\quad \mathbf{if}\;\; Med({\mathcal{P}}_{k}^{\prime})< Count({p}_{i}),
		\\
		&{ \quad\;\; 1.0 \;\;\;\quad } ,  \quad \mathbf{if}\;\; Med({\mathcal{P}}_{k}^{\prime}) \geq Count({p}_{i}).
	\end{aligned}
	\right.
\end{equation}

By employing the above strategy, each classifier would be expert in distinguishing the predicates, particularly in the newly-added group. For example, since we would under-sample the instances in Group 3 for training the $4^{\rm th}$ and $5^{\rm th}$ classifiers, the $3^{\rm rd}$ classifier is more likely to achieve a better performance in distinguishing the predicates in Group 3, as we retain all the samples of this group to let the $3^{\rm rd}$ classifier adequately learn the discriminating representations.

\textbf{Class Probability Prediction} aims to parse the sampled instances into the class probability logits for further loss computation and model optimization. For an object pair $(o_{i}, o_{j})$ chosen by the Median Re-Sampling strategy, after obtaining the subject feature $\mathbf{x}^{\prime}_i$, the object feature $\mathbf{x}^{\prime}_j$, and their union feature $\mathbf{u}_{ij}$, the class probability prediction $\mathbf{w}_{ij}^{k} $ generated by the classifier $\mathcal{C}_{k}$ is calculated as follows:
\begin{equation}
    \mathbf{w}_{ij}^{k} = {\rm Softmax}(FC([\mathbf{x}^{\prime}_i , \mathbf{x}^{\prime}_j]) \otimes \mathbf{u}_{ij}),
\end{equation}
where $\otimes$ denotes the element-wise product.

\textbf{Parallel Classifier Optimization} aims to regularize the final classifier $\mathcal{C}_{K}$ by jointly optimizing all the classifiers. In the training stage, the parameters of all the $K$ predicate classifiers would be optimized simultaneously, where the objective function can be defined as:
\begin{equation}
	\mathcal{L}_{PCO}=\sum_{k=1}^{K} \frac{1}{|\mathcal{D}_k|}\sum_{{(o_i,o_j)} \in \mathcal{D}_k}  \mathcal{L}_{CE} ({y}_{ij}, \mathbf{w}_{ij}^{k}),
\end{equation}
where $\mathcal{D}_k$ denotes the set of the object pairs chosen by the Median Re-Sampling strategy, $| \cdot |$ denotes the length of the given set, ${y}_{ij}$ denotes the ground-truth predicate label of the object pair $(o_{i}, o_{j})$, and $\mathcal{L}_{CE}(\cdot)$ is a regular Cross-Entropy cost function. 

The Parallel Classifier Optimization can be seen as a ``weak constraint'' for Group Collaborative Learning, since we expect that gathering gradients from all the classifiers would facilitate the convergence of the final classifier $\mathcal{C}_{K}$.

\textbf{Collaborative Knowledge Distillation} aims to establish a knowledge transfer mechanism to promote the unbiased prediction capability of the final classifier $\mathcal{C}_{K}$. As aforementioned, each classifier specializes in distinguishing the predicates, particularly within the newly-added group. In order to preserve and translate this well-learned knowledge to compensate for the under-fitting on the head classes, we propose the Collaborative Knowledge Distillation (CKD), whose objective function is defined as:
\begin{equation}
	\mathcal{L}_{CKD}=\frac{1}{| \mathcal{Q} |} \sum_{{(m, n)} \in \mathcal{Q}}   \frac{1}{|\mathcal{D}_n|}\sum_{{(o_i,o_j)} \in \mathcal{D}_n}  \mathcal{L}_{KL} (\mathbf{w}_{ij}^{m}, \widehat{\mathbf{w}}_{ij}^{{n}}),
\end{equation}
where $\mathcal{Q}$ denotes the set of pairwise knowledge matching from the classifier $\mathcal{C}_{m}$ to the classifier $\mathcal{C}_{n}$ ($m<n$). We provide two alternatives, namely Adjacent and Top-Down strategy, to configure the set $\mathcal{Q}$ (these two strategies are illustrated in Figure \ref{parameters} and Parameter Analysis). Note that the output $\mathbf{w}_{ij}^{n}$ generated by the classifier $\mathcal{C}_{n}$ incorporates new predicate classes which are not included in the previous classification space $\mathcal{P}_{m}^{\prime}$, we utilize $\widehat{\mathbf{w}}_{ij}^{{n}}$ to indicate the sliced output by cutting off the incrementally-added classes which are not included in $\mathcal{P}_{m}^{\prime}$, thus ensuring that $\widehat{\mathbf{w}}_{ij}^{{n}}$ shares the same dimension as $\mathbf{w}_{ij}^{m}$. $\mathcal{L}_{KL}(\cdot)$ is a regular Kullback-Leibler Divergence loss, which is defined as:

\begin{equation}
	\mathcal{L}_{KL} (\mathbf{w}_{m}, \widehat{\mathbf{w}}_{n}) = - \sum_{l=1}^{L} \mathbf{w}_{m}^{l} \log \widehat{\mathbf{w}}_{n}^{l}.
\end{equation}

By taking the previous predicate probability output $\mathbf{w}_{ij}^{m}$ from the classifier $\mathcal{C}_{m}$ as the soft label, CKD forces the current classifier $\mathcal{C}_{n}$ to mimic the prediction behaviour that $\mathcal{C}_{m}$ is expert in, thus can be treated as a ``strong constraint''.

Ultimately, the objective function of our proposed Group Collaborative Learning (GCL) is the combination of PCO and CKD, which is defined as:
\begin{equation}
	\mathcal{L}_{GCL} = \mathcal{L}_{PCO} + \alpha\mathcal{L}_{CKD},
\end{equation}
where $\alpha$ is the pre-defined hyper-parameters to weigh the total loss $\mathcal{L}_{GCL}$. By employing these two types of constraint, we effectively mitigate the overwhelming punishments to the tail classes and compensate for the under-fitting on the head ones, which benefits in establishing a reasonable trade-off during the predicate predictions.


\begin{table*}[t]
	\small
	\vspace{-0.4cm}
	\begin{tabular}{p{2.9cm}|p{1.1cm}<{\centering}p{1.1cm}<{\centering}p{1.1cm}<{\centering}|p{1.1cm}<{\centering}p{1.1cm}<{\centering}p{1.1cm}<{\centering}|p{1.1cm}<{\centering}p{1.1cm}<{\centering}p{1.1cm}<{\centering}}
		\hline
		\multicolumn{1}{c|}{\multirow{2}{*}{Model}} & \multicolumn{3}{c|}{PredCls}& \multicolumn{3}{c|}{SGCls}& \multicolumn{3}{c}{SGDet}\\ \cline{2-10} 
		\multicolumn{1}{c|}{}& \multicolumn{1}{c}{mR@20} & \multicolumn{1}{c}{mR@50} & \multicolumn{1}{c|}{mR@100} & \multicolumn{1}{c}{mR@20} & \multicolumn{1}{c}{mR@50} & \multicolumn{1}{c|}{mR@100} & \multicolumn{1}{c}{mR@20} & \multicolumn{1}{c}{mR@50} & \multicolumn{1}{c}{mR@100} \\ \hline
		
		IMP+$^\dag$ & -  &9.8  &10.5 &-  &5.8  &6.0 &-  & 3.8 &4.8 \\
		KERN$^\dag$ & -  &17.7  &19.2 &-  &9.4  &10.0 &-  & 6.4 &7.3 \\
		GPS-Net$^\dag$ & 17.4  &21.3  &22.8 &10.0  &11.8  &12.6 &6.9  & 8.7 &9.8 \\
		PCPL$^\dag$ & -  &35.2  &37.8 &-  &18.6  &19.6 &-  & 9.5 &11.7 \\
		VTransE+ &13.6  &17.1  &18.6 &6.6  &8.2  &8.7 &5.1  & 6.8 &8.0 \\
	    SG-CogTree & 22.9  &28.4  &31.0 &13.0  &15.7  &16.7 &7.9  & 11.1 &12.7 \\
		BGNN & -  &30.4  &32.9 &-  &14.3  &16.5 &-  & 10.7 &12.6 \\
		
		 \hline
		
		Motifs &11.7  &14.8  &16.1 &6.7  &8.3 &8.8 &5.0  &6.8  & 7.9\\ 
		Motifs + Reweight$_{d}$ &14.3  &17.3  &18.6 &9.5  &11.2 &11.7 &6.7  &9.2  & 10.9\\
		Motifs + TDE$_{d}$ &18.5  &25.5  &29.1 &9.8  &13.1 &14.9 &5.8  &8.2  & 9.8\\
		Motifs + CogTree$_{d}$ &20.9  &26.4  &29.0 &12.1  &14.9 &16.1 &7.9  &10.4  & 11.8\\
		Motifs + DLFE$_{d}$ &22.1  &26.9  &28.8 &12.8  &15.2 &15.9 &8.6  &11.7  & 13.8\\
		Motifs + EBM$_{d}$ &14.2  &18.0  &28.8 &8.2  &10.2 &11.0 &5.7  &7.7  & 9.3\\
		\textbf{Motifs + GCL} &\textbf{30.5}  &\textbf{36.1}  &\textbf{38.2} &\textbf{18.0}  &\textbf{20.8}  &\textbf{21.8}  &\textbf{12.9}  &\textbf{16.8}  &\textbf{19.3} \\
		\hline
		
		VCTree &13.1 &16.7 &18.1 & 9.6  & 11.8  &12.5 &5.4  &7.4 &8.7 \\
		VCTree + Reweight$_{d}$&16.3 &19.4 &20.4 & 10.6  & 12.5  &13.1 &6.6  &8.7 &10.1 \\
		VCTree + TDE$_{d}$&18.4 &25.4 &28.7 & 8.9  & 12.2  &14.0 &6.9  &9.3 &11.1 \\
		VCTree + CogTree$_{d}$&22.0 &27.6 &29.7 & 15.4  & 18.8  &19.9 &7.8  &10.4 &12.1 \\
		VCTree + DLFE$_{d}$&20.8 &25.3 &27.1 & 15.8  & 18.9  &20.0 &8.6  &11.8 &13.8 \\
		VCTree + EBM$_{d}$&14.2 &18.2 &19.7 & 10.4  & 12.5  &13.5 &5.7  &7.7 &9.1 \\
		\textbf{VCTree + GCL}  &\textbf{31.4}   &\textbf{37.1}  &\textbf{39.1}  &\textbf{19.5}  &\textbf{22.5}  &\textbf{23.5}  &\textbf{11.9}  &\textbf{15.2} &\textbf{17.5} \\
		 \hline 
		 
		SHA &14.4  &18.8  &20.5 &8.7  &10.9 &11.6 &5.7  &7.8  & 9.1\\
		
		\textbf{SHA + GCL (ours)} &\textbf{35.6}  &\textbf{41.6}  &\textbf{44.0}  &\textbf{19.6}  &\textbf{23.0}  &\textbf{24.3}  &\textbf{14.2}  &\textbf{17.9}  &\textbf{20.9} \\
		 \hline
		
	\end{tabular}
\vspace{0.02cm}
\caption{Performance comparison of different methods on PredCls, SGCls, and SGDet tasks of VG150 with respect to mR@20/50/100 (\%). The superscript $\dag$ denotes that the method employs Faster R-CNN with VGG-16 as the object detector, while the subscript $d$ denotes that the method is model-agnostic and targets to address the biased relationship predictions in SGG.}
\vspace{-0.4cm}
\label{result_VG}
\end{table*}

\begin{table}[t]
	\small
	\vspace{0.0cm}
	\begin{tabular}{p{2.3cm}|p{1.7cm}<{\centering}|p{1.4cm}<{\centering}|p{1.4cm}<{\centering}}
		\hline
		\multicolumn{1}{c|}{\multirow{2}{*}{Model}} & \multicolumn{1}{c|}{PredCls}& \multicolumn{1}{c|}{SGCls}&\multicolumn{1}{c}{SGDet}
		\\ 
		\cline{2-4} 
		\multicolumn{1}{c|}{} & \multicolumn{1}{c|}{mR 50/100} & \multicolumn{1}{c|}{mR 50/100} & \multicolumn{1}{c}{mR 50/100} \\ \hline
		
		VTransE &14.0 / 15.0  & 8.1 / 8.7 & 5.8 / 6.6  \\ 
		\textbf{VTransE + GCL} &\textbf{30.4} / \textbf{32.3} &\textbf{16.6} / \textbf{17.4}  &\textbf{14.7} / \textbf{16.4}  \\
		\hline
		
		Motifs &16.4 / 17.1  &8.2 / 8.6  &6.4 / 7.7 \\ 
		\textbf{Motifs + GCL} &\textbf{36.7} / \textbf{38.1} &\textbf{17.3} / \textbf{18.1}  &\textbf{16.8} / \textbf{18.8}  \\
		\hline
		
		VCTree &16.6 / 17.4 & 7.9 / 8.3  & 6.5 / 7.4  \\
		\textbf{VCTree + GCL} &\textbf{35.4} / \textbf{36.7} &\textbf{17.3} / \textbf{18.0} &\textbf{15.6} / \textbf{17.8}\\
		 \hline
		 
		SHA &19.5 / 21.1  & 8.5 / 9.0  & 6.6 / 7.8    \\
		\textbf{SHA + GCL} &\textbf{41.0} / \textbf{42.7} &\textbf{20.6} / \textbf{21.3}  &\textbf{17.8} / \textbf{20.1}  \\
		 \hline
		
	\end{tabular}
\vspace{0.02cm}
\caption{Performance comparison of different methods on three tasks of GQA200 with respect to mR@50/100 (\%).}
\vspace{-0.4cm}
\label{result_GQA}
\end{table}

\section{Experiments}

\subsection{Experimental Settings}
\noindent\textbf{Dataset.}
We present the experimental results on two datasets: Visual Genome (VG)~\cite{krishna2016visual} and GQA~\cite{hudson2019gqa}. VG is the most widely-used benchmark for SGG, which is composed of more than 108K images and 2.3M relation instances. Following the prior approaches \cite{xu2017scene, chen2019knowledge, lin2020gps, yan2020pcpl, zhang2017visual, li2021bipartite, zellers2018neural, tang2019learning, tang2020unbiased, yu2020cogtree, chiou2021recovering, suhail2021energy}, we adopt the most widely-used VG150 split, which contains the most frequent 150 object classes and 50 predicate classes. GQA is another vision-and-language benchmark with more than 3.8M relation annotations. In order to achieve a representative split like VG150, we manually clean up a substantial fraction of annotations that have poor-quality or ambiguous meanings, and then select Top-200 object classes as well as Top-100 predicate classes by their frequency, thus establishing the GQA200 split. For both VG150 and GQA200, we use 70$\%$ of the images for training and the remaining 30$\%$ for testing. We also follow~\cite{zellers2018neural} to sample a 5K validation set from the training set for parameter tuning.

\noindent\textbf{Tasks.}
To comprehensively evaluate the performance, we follow three conventional tasks: 1) Predicate Classification (\textbf{PredCls}) predicts the relationships of all the pairwise objects by employing the given ground-truth bounding boxes and classes; 2) Scene Graph Classification (\textbf{SGCls}) predicts the objects classes and their pairwise relationships by employing the given ground-truth object bounding boxes; and 3) Scene Graph Detection (\textbf{SGDet}) detects all the objects in an image, and predicts their bounding boxes, classes and pairwise relationships.

\noindent\textbf{Evaluation Metrics.} Following \cite{yan2020pcpl, lin2020gps, li2021bipartite, tang2020unbiased, yu2020cogtree, chiou2021recovering, suhail2021energy}, we use mean Recall@K
(mR@K)\cite{tang2019learning,chen2019knowledge}, which computes the average Recall@K (R@K) for each predicate class, to evaluate the unbiased SGG. As R@K is easily dominated by the head classes due to the extremely unbiased dataset, mR@K could give a fair performance appraisal for both head and tail classes, which is widely used as an unbiased evaluation metric.

\noindent\textbf{Implementation Details.} We adopt a pre-trained Faster R-CNN \cite{Ren2017Faster} with ResNeXt-101-FPN\cite{xie2017aggregated} provided by \cite{tang2020unbiased} as the object detector. We employ Glove\cite{pennington2014glove} to obtain the semantic embedding. The object encoder and the relation encoder contain four and two SHA layers, respectively. We set the division threshold $\mu=4$, and employ the Top-Down strategy (each classifier is forced to learn the prediction behavior from all its predecessors, see Figure \ref{parameters} for more details) to construct the pairwise knowledge matching set $\mathcal{Q}$. The hyper-parameter $\alpha$ which balances the optimization objective is set to be 1.0. We optimize the proposed network by the Adam optimizer with a momentum of 0.9. For all three tasks, the total training stage lasts for 60,000 steps with a batch size of 8. The initial learning rate is 0.001, and we adopt the same warm-up and decayed strategy as \cite{tang2020unbiased}. One RTX2080 Ti is used to conduct all the experiments.

\subsection{Compared Methods}
We want to declare that our proposed method is not only powerful in generating unbiased scene graphs, but also applicable for a variety of SGG approaches. For the former, we compare it with state-of-the-art approaches, including re-produced IMP+\cite{xu2017scene}, KERN\cite{chen2019knowledge}, GPS-Net\cite{lin2020gps}, PCPL\cite{yan2020pcpl}, re-produced VTransE+\cite{zhang2017visual} and BGNN\cite{li2021bipartite}. For the latter, we adopt two typical baselines, namely Motifs\cite{zellers2018neural} and VCTree\cite{tang2019learning}, to give a fair comparison with other model-agnostic approaches, such as Reweighting\cite{chiou2021recovering},  TDE\cite{tang2020unbiased}, CogTree\cite{yu2020cogtree}, DLFE\cite{chiou2021recovering} and EBM\cite{suhail2021energy}. 

Table \ref{result_VG} and Table \ref{result_GQA} present the performance of different approaches conducted on VG150 and GQA200, respectively. We have several observations as follows: 1) Our proposed SHA+GCL significantly outperforms all the baselines on all three tasks. To the best of our knowledge, our work is the first to breakthrough the 40\% precision in both mR@50 and mR@100 on PredCls, and we also achieve the best performance on SGCls and SGDet. 2) Motifs+GCL and VCTree+GCL nearly double the performance in mean Recall on all three tasks compared with Motifs and VCTree. It demonstrates that the proposed GCL is model-agnostic and can largely enhance the unbiased relationship predictions. 3) Compared with Motifs+GCL and VCTree+GCL, we witness an obvious performance gain in SHA+GCL. It indicates that the proposed SHA module could facilitate both the intra-modal refinement and the inter-modal interaction, thus leading to more accurate predictions. In conclusion, SHA+GCL effectively addresses two aforementioned concerns in SGG, \ie, insufficient modality fusion and biased relationship predictions.

\subsection{Ablation Study}
As aforementioned, we propose the Stacked Hybrid Attention (SHA) network to improve the object encoder and the relation encoder, and propose the Group Collaborative Learning (GCL) strategy, which employs the Parallel Classifier Optimization (PCO) as the ``weak constraint'' and Collaborative Knowledge Distillation (CKD) as the ``strong constraint'', to guide the training of the decoder. In order to prove the effectiveness of the above components, we test various ablation models on VG150 as follows: 

\vspace{-0.2cm}
\begin{itemize}
\setlength{\itemsep}{0pt}
\setlength{\parsep}{0pt}
\setlength{\parskip}{0pt}
	\item w/o-GCL: To evaluate the effectiveness of GCL, we let the relation decoder be a one-layer classifier, where a regular Cross-Entropy loss is performed.
	
	\item w/o PCO\&CKD: To evaluate the effectiveness of PCO in GCL, we remove the PCO loss and CKD loss, and only employ the Median Re-Sampling strategy and a regular Cross-Entropy loss in the optimization step.
	
	\item w/o CKD: To evaluate the effectiveness of CKD in GCL, we remove the CKD loss but retain all the classifiers to compute the PCO loss.
	
	\item w/o CA or w/o SA: To evaluate the effectiveness of SHA, we remove either the Cross-Attention (CA) unit or the Self-Attention (SA) unit in every SHA layer.

\end{itemize}
\vspace{-0.2cm}

\begin{table}[t]
	\small
	\vspace{0.4cm}
	\begin{tabular}{p{2.3cm}|p{1.4cm}<{\centering}|p{1.4cm}<{\centering}|p{1.4cm}<{\centering}}
		\hline
		\multicolumn{1}{c|}{\multirow{2}{*}{Model}} & \multicolumn{1}{c|}{PredCls}& \multicolumn{1}{c|}{SGCls}&\multicolumn{1}{c}{SGDet}
		\\ 
		\cline{2-4} 
		\multicolumn{1}{c|}{} & \multicolumn{1}{c|}{mR 50/100} & \multicolumn{1}{c|}{mR 50/100} & \multicolumn{1}{c}{mR 50/100} \\ \hline
		
		w/o - GCL &18.8 / 20.5  &10.9 / 11.6  &7.8 / 9.1  \\ 
		w/o - PCO\&CKD & 35.2 / 37.4   & 20.1 / 21.2  & 14.6 / 16.9   \\
		w/o - CKD &39.3 / 41.7  &22.0 / 23.2  &16.5 / 19.0   \\
		w/o - CA &39.8 / 42.5   &22.6 / 23.6  &16.8 / 19.3   \\
		w/o - SA &39.2 / 41.5   &22.6 / 23.7  &17.5 / 20.1   \\
		
		\hline
		
		SHA + GCL &\textbf{41.6} / \textbf{44.0} &\textbf{23.0} / \textbf{24.3}  &\textbf{17.9} / \textbf{20.9}  \\
		\hline
	\end{tabular}
\vspace{0.02cm}
\caption{Ablation study of the proposed method on VG150.}
\vspace{-0.2cm}
\label{result_AB}
\end{table}

\begin{figure}
	\centering
	\begin{subfigure}{1\linewidth}
		\includegraphics[width=1.0\textwidth]{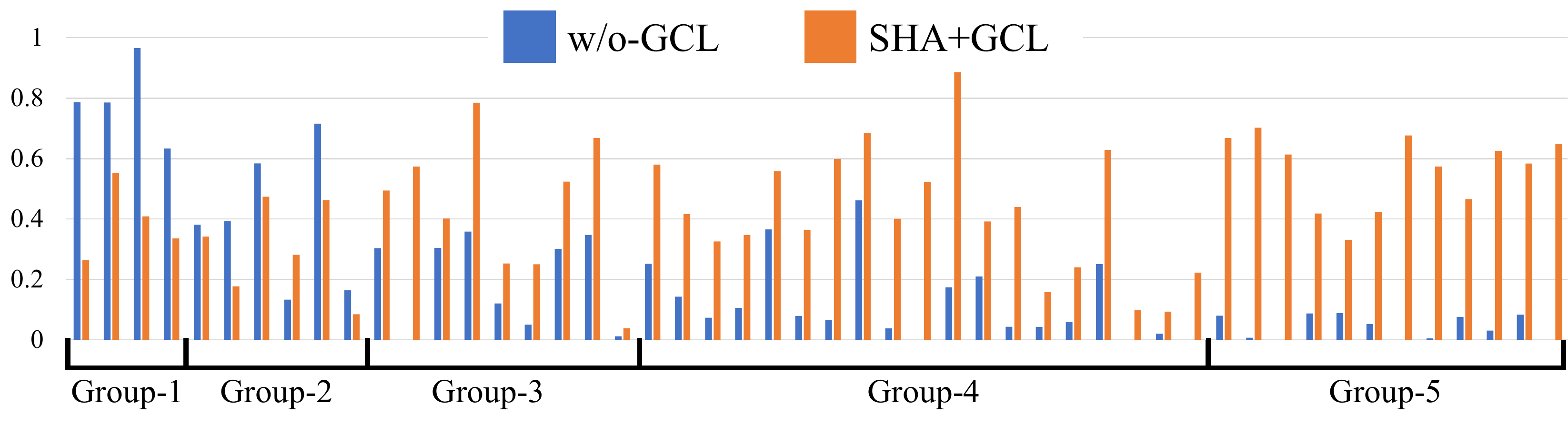}
		\caption{\scriptsize{R@100 of all the predicate classes of w/o-GCL and SHA+GCL on VG150.}}
		\label{norVSgist}
	\end{subfigure}
 	\begin{subfigure}{1\linewidth}
		\includegraphics[width=1.0\textwidth]{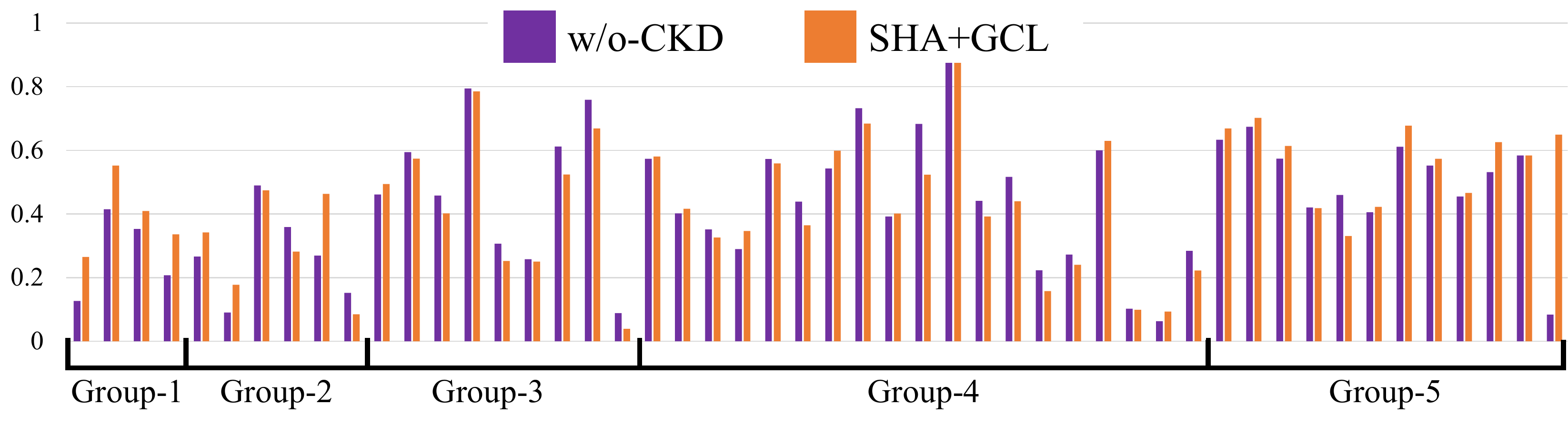}
		\caption{\scriptsize{R@100 of all the predicate classes of w/o-CKD and SHA+GCL on VG150.}}
		\label{gclVSgist}
	\end{subfigure}
	\vspace{-0.2cm}
	\caption{R@100 of 50 predicate classes on PredCls on VG150.}
	\vspace{-0.4cm}
	\label{ablation}
\end{figure}

Table \ref{result_AB} presents the results of all the ablation models. We have several observations as follows: 1) Compared with w/o-GCL, SHA+GCL nearly doubles the performance. Moreover, in Figure \ref{norVSgist}, we compare w/o-GCL and SHA+GCL with respect to R@100 of all the predicate classes. As can be observed, SHA+GCL obviously improves the performance on most of the predicate classes, only with an acceptable decay on the head classes in Group 1 and Group 2, showing a powerful capability in generating unbiased scene graphs. 2) Compared with w/o-PCO\&CKD, w/o-CKD evidently improves the prediction performance, demonstrating that the ``weak constraint'', namely gathering gradients from all the classifiers, would facilitate the convergence of the final classifier. 3) Compared with w/o-CKD, we witness an obvious performance gain in SHA+GCL. Moreover, we compare w/o-CKD and SHA+GCL on the detailed precision towards every predicate class on VG150. As shown in Figure \ref{gclVSgist}, CKD effectively prevents the model from sacrificing much on the head classes, as well as achieves a comparable performance towards the tail predictions. It demonstrates that the ``strong constraint'', namely a knowledge transfer paradigm, could effectively compensate for the under-fitting on the head classes by preserving the discriminating capability learned previously, and thus benefits in achieving a reasonable trade-off. 4) From the last three rows in Table \ref{result_AB}, we witness an obvious performance decay when removing either the CA unit or the SA unit. It verifies that combining both attentions would effectively alleviate the insufficient modality fusion, thus leading to more accurate predictions.

\subsection{Parameter Analysis}

\begin{figure}[t]
	\centering
	\includegraphics[width=0.49\textwidth]{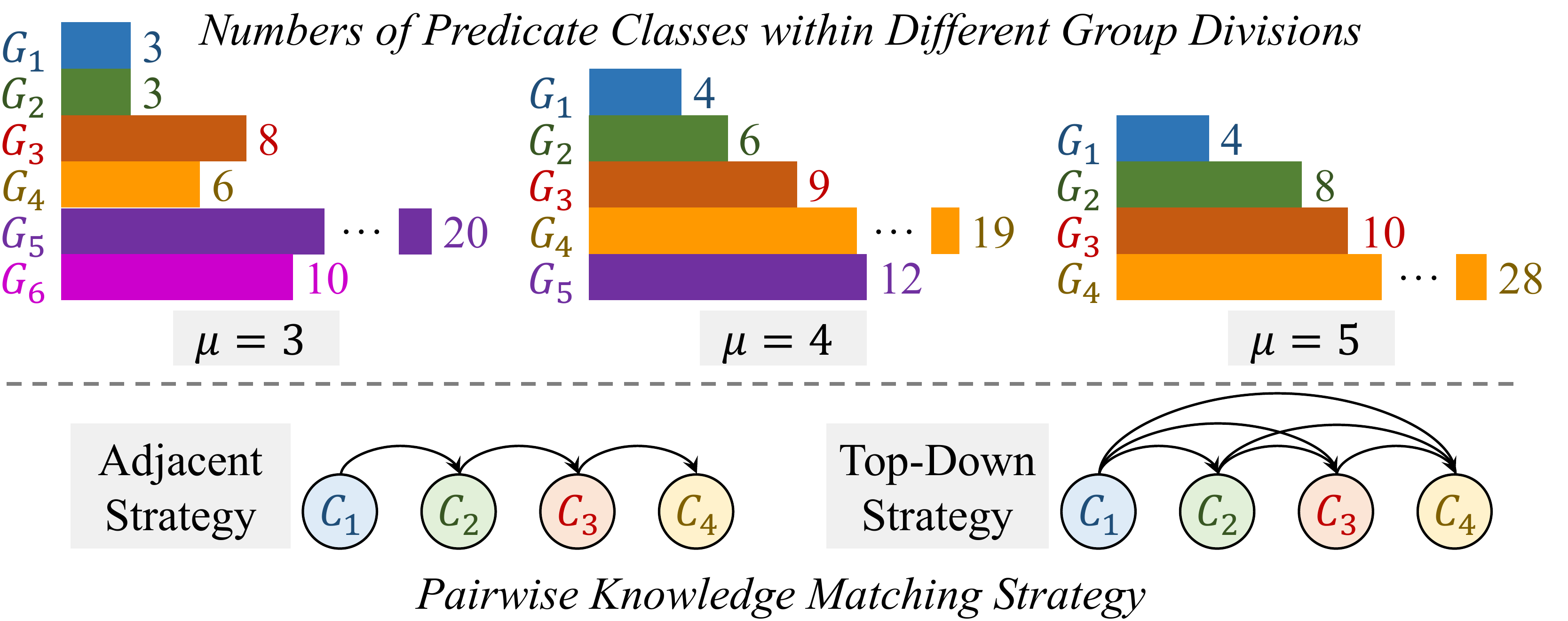}
	\vspace{-0.5cm}
	\caption{Illustration of three configurations of the balanced group divisions according to the threshold $\mu$ (top), and two alternatives of the pairwise knowledge matching strategy (down).}
	\vspace{-0.4cm}
	\label{parameters}
\end{figure}

As aforementioned, the threshold $\mu$ and the organization strategy would influence the performance of GCL. As Figure \ref{parameters} illustrates, for the former, we set $\mu=$3, 4, and 5, and obtain 6, 5, and 4 group divisions, respectively. For the latter, we provide two alternatives, namely Adjacent and Top-Down strategy, whose difference is whether each classifier could learn the knowledge from its nearest predecessor (Adjacent) or from all the predecessors (Top-Down).

\begin{table}
	\small
	\vspace{0.4cm}
	\begin{tabular}{p{0.4cm}<{\centering}|p{1.4cm}<{\centering}|p{1.4cm}<{\centering}|p{1.4cm}<{\centering}|p{1.4cm}<{\centering}}
		\hline
		\multicolumn{2}{c|}{Model} & \multicolumn{1}{c|}{PredCls}& \multicolumn{1}{c|}{SGCls}&\multicolumn{1}{c}{SGDet}
		\\ 
		
		\hline
		\multicolumn{1}{c|}{$\mu$} & \multicolumn{1}{c|}{Strategy} & \multicolumn{1}{c|}{mR 50/100} & \multicolumn{1}{c|}{mR 50/100} & \multicolumn{1}{c}{mR 50/100} \\ \hline
		3 & Adjacent &40.0 / 42.4 & 22.5 / 23.4  & 16.8 / 19.2    \\
		4 & Adjacent &41.0 / 43.5 &23.0 / 23.9  &17.3 / 19.7   \\
		5 & Adjacent &39.4 / 41.7   & 21.8 / 23.0   & 16.7 / 19.1    \\\hline
		3 & Top-down &40.9 / 43.2   &22.9 / 23.8  &17.0 / 19.9   \\
		4 & Top-down &\textbf{41.6} / \textbf{44.0} &\textbf{23.0} / \textbf{24.3}  &\textbf{17.9} / \textbf{20.9}  \\
		5 & Top-down &39.7 / 42.0   &23.1 / 23.8  &16.9 / 19.6   \\
		 \hline
	\end{tabular}
\vspace{0.02cm}
\caption{Parameter analysis towards the threshold $\mu$ and the pairwise knowledge matching strategies of GCL on VG150.}
\vspace{-0.2cm}
\label{result_PA}
\end{table}

Table \ref{result_PA} presents the performance comparisons, where $\mu=4$ and the Top-Down strategy is the best combination.


\section{Conclusion}


In this work, we declare two concerns that restrict the practical applications of SGG, namely insufficient modality fusion and biased relationship predictions. To address such deficiency, we propose the Stacked Hybrid-Attention network and the Group Collaborative Learning strategy. In this way, we establish a new state-of-the-art in the unbiased metric and provide a model-agnostic debiasing method. In the future, we plan to explore more robust group dividing methods and devise more knowledge distillation strategies.

\setlength{\parskip}{8pt}
\noindent\textbf{Acknowledgments.} This work is supported by the National Natural Science Foundation of China, No.: 62176137, No.:U1936203, and No.: 62006140; the Shandong Provincial Natural Science and Foundation, No.: ZR2020QF106; Beijing Academy of Artificial Intelligence(BAAI); Ant Group.
\setlength{\parskip}{0pt}

\newpage
{\small
\bibliographystyle{ieee_fullname}
\bibliography{egbib}
}

\newpage

\setcounter{section}{0}

\setlength{\parskip}{0pt}
\section{Introduction}

In this supplementary material, we present more analyses, experiments, and visualization results, as well as discuss the limitations and future work of our method.

\section{Parameter Statistics}
We compare the total number of parameters between three baseline methods (\ie, Motifs, VCTree, and SHA) and their enhanced versions that are equipped with our model-agnostic GCL in Table \ref{para_counts}. As can be observed, compared with the original methods which possess a massive number of total trainable parameters (about 200M), GCL only additionally introduces a limited number of parameters (about 2M), which could hardly influence the overall training procedure.

\section{Detailed Performance}
We present the complete results of our experiments employing the regular Recall@K\cite{lu2016visual}, the unbiased Mean Recall@K\cite{tang2019learning,chen2019knowledge}, and their mean\cite{lin2020gps} on all three tasks (\ie, PredCls, SGCls, and SGDet) on VG150\cite{krishna2016visual} and GQA200\cite{hudson2019gqa} dataset in Table \ref{result_all}, where $K \in \{50,100\}$. Note that all the methods are implemented with a pre-trained Faster R-CNN\cite{Ren2017Faster} with ResNeXt-101-FPN\cite{xie2017aggregated} provided by \cite{tang2020unbiased} as the object detector, thus we could give a fair comparison to prove the superiority of our method.

From Table \ref{result_all}, we observe that 1) our proposed SHA+GCL achieves the best performance on all three tasks towards the unbiased metric mR@K in both two datasets. In VG150, we breakthrough the 40\% precision in both mR@50 and mR@100 on PredCls, and 20\% precision in mR@100 on both SGCls and SGDet, thus establishing a new state-of-the-art in the unbiased metric. 2) Our improvement towards the relation decoder, namely GCL strategy, is model-agnostic and could largely enhance the unbiased SGG. In both VG150 or GQA200, the method equipped with GCL nearly doubles the performance compared with the original one, showing the outstanding capability in generating unbiased scene graphs.

\begin{figure}[t]
	\centering
	\includegraphics[width=0.49\textwidth]{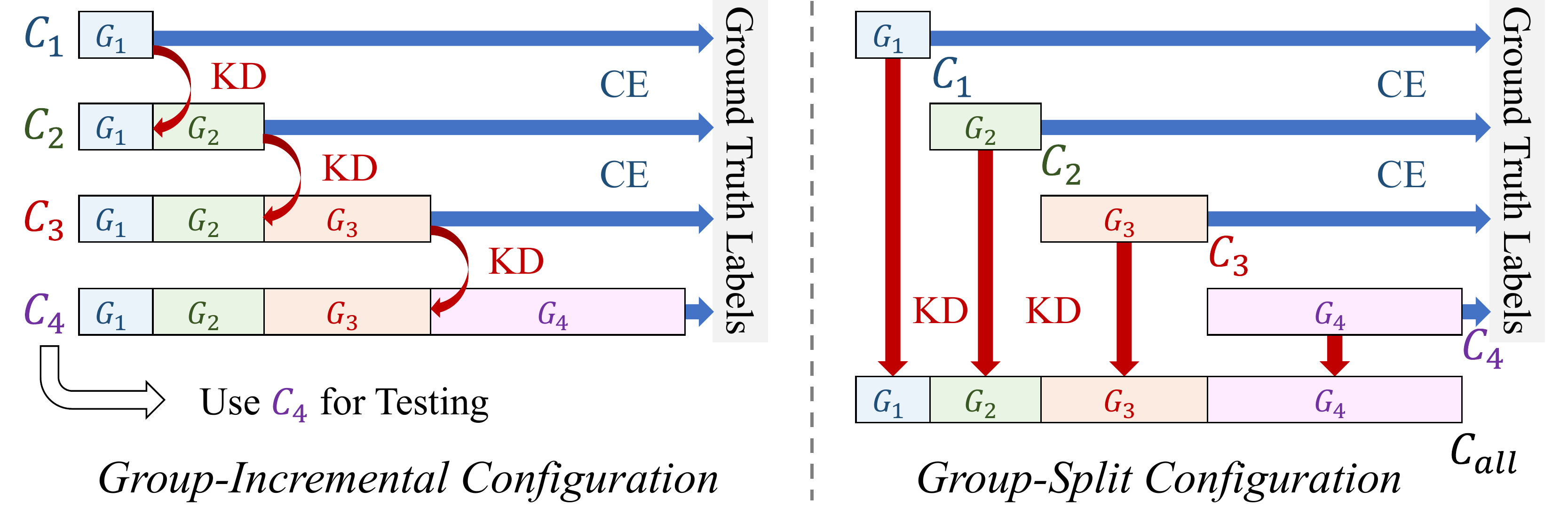}
	\vspace{-0.4cm}
	\caption{The group-incremental configuration (left) may not be the only alternative to fulfill the ``conquer'' step in GCL. For example, the group-split configuration (right) is another promising strategy. Therefore, we aim to explore more robust group dividing methods and classifier configuration strategies in the future.}
	\vspace{-0.4cm}
	\label{future_work}
\end{figure}

\section{Visualization Results}
\begin{figure*}[t]
	\begin{center}
		\includegraphics[width=1.0\textwidth]{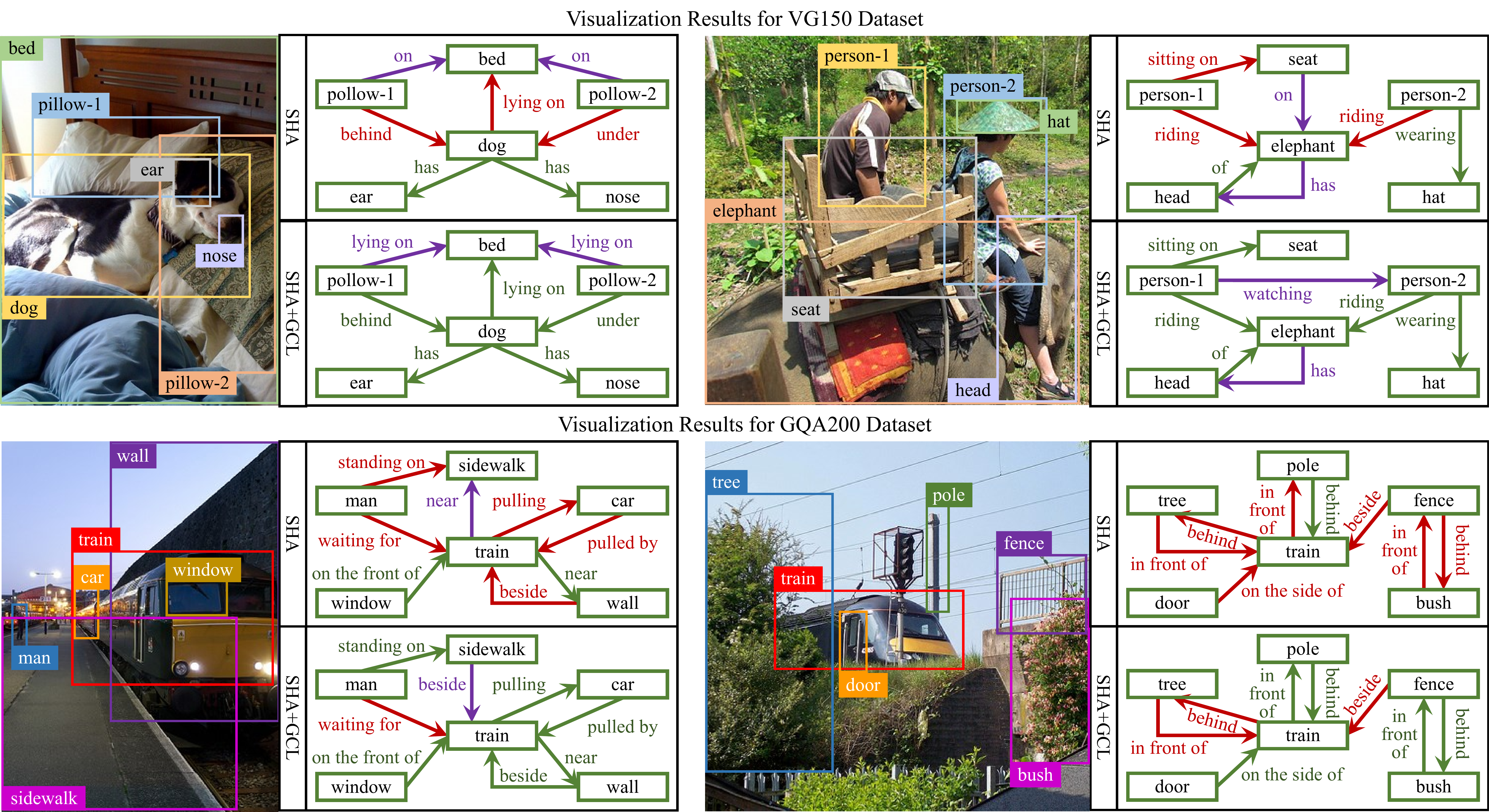}
	\end{center}
	\vspace{-0.4cm}
	\caption{Qualitative comparisons between SHA and SHA+GCL with regard to R@20 on PredCls setting. Green edges represent the ground truth relationships that are correctly predicted, red edges represent the ground truth relationships that are failed to be detected, and purple edges represent the reasonable relationships which are predicted by the model but are not annotated in the ground truth.}
	\vspace{-0.4cm}
	\label{visualization}
\end{figure*}

To get an intuitive perception of the superior performance in generating unbiased scene graphs of our proposed GCL, we visualize several PredCls examples generated from the biased SHA and the unbiased SHA+GCL. As shown in Figure \ref{visualization}, the model employing the proposed GCL strategy prefers to providing more informative and specific relationship predictions (\eg, lying on and riding) rather than common and trivial ones (\eg, on and has), \eg, ``person1-riding-elephant'' in the top-right example and ``train-pulling-car'' in the bottom-left example. Moreover, the model equipped with our model-agnostic GCL could also capture potential reasonable relationships, such as ``person1-watching-person2'' in the top-right example and ``sidewalk-beside-train'' in the bottom-left example. In a nutshell, the proposed GCL could enhance the unbiased relationship predictions, thus achieving more informative scene graphs to support various down-stream tasks.

\section{Limitations and Future Work}

In this section, we would like to discuss the limitations of our method, based on which we provide several potential directions to further improve our SHA+GCL.

\subsection{More Configurations Could be Further Explored}
As aforementioned, we follow the intuition of ``divide-conquer-cooperate'' to address the biased relationship predictions. In the second step, namely ``conquer'', we borrow the idea from class-incremental learning \cite{hu2020learning} and employ the group-incremental configuration. Actually, we employ this configuration mainly due to its simplicity and efficiency, as we could directly leverage the final classifier that covers all the candidate classes to obtain the predictions in the evaluation stage. However, we should argue that it is not the only alternative to fulfill the ``conquer'' step. Therefore, in the future, we aim to explore more robust group dividing methods as well as classifier configuration strategies to promote the unbiased SGG, \eg, the group-split configuration in Figure \ref{future_work}.

\subsection{``Strong Constraint'' Could be Further Enhanced}
As aforementioned, in the ``cooperate'' step, we use the collaborative knowledge distillation to establish an effective knowledge transfer mechanism, where a regular Kullback-Leibler Divergence loss is employed. However, since various novel methods have been proposed in the knowledge distillation area, we could further enhance our GCL by devising more efficient strategies, thus strengthening the ``Strong Constraint'' and promoting the unbiased SGG.

\begin{table*}[t]
	\small
	\begin{tabular}{p{2.0cm}<{\centering}p{1.1cm}<{\centering}p{1.1cm}<{\centering}|p{2.0cm}<{\centering}p{1.1cm}<{\centering}p{1.1cm}<{\centering}|p{2.0cm}<{\centering}|p{1.1cm}<{\centering}p{1.1cm}}
		\hline
		\multicolumn{1}{c}{Model}& \multicolumn{1}{c}{Fixed} & \multicolumn{1}{c|}{Trainable} & \multicolumn{1}{c}{Model} & \multicolumn{1}{c}{Fixed} & \multicolumn{1}{c|}{Trainable} & \multicolumn{1}{c}{Model} & \multicolumn{1}{c}{Fixed} & \multicolumn{1}{c}{Trainable}\\ \hline
		
		Motifs  & 158.7M & 208.5M & VCTree & 158.7M & 199.8M & SHA & 158.7M & 228.8M \\ 
		Motifs + GCL  & 158.7M & 210.5M & VCTree + GCL & 158.7M & 201.8M & SHA + GCL & 158.7M & 230.9M \\  \hline
		
	\end{tabular}
	\caption{Comparison of different methods on the number of parameters.       ``Fixed'' counts the number of parameters that belong to the pre-trained object detector, and ``Trainable'' counts the number of parameters that can be updated during the training procedure.}
	\label{para_counts}
	\vspace{0.1cm}
\end{table*}

\begin{table*}[t]
	\small
	\begin{tabular}{p{3.2cm}|p{1.4cm}<{\centering}p{1.4cm}<{\centering}|p{1.4cm}<{\centering}p{1.4cm}<{\centering}|p{1.4cm}<{\centering}p{1.4cm}<{\centering}|p{0.5cm}<{\centering}p{0.5cm}}
		\hline
		\multicolumn{9}{c}{\multirow{2}{*}{Evaluation on Visual Genome Dataset}}\\
		\multicolumn{9}{c}{}\\
		\hline
		\multicolumn{1}{c|}{\multirow{2}{*}{Model}} & \multicolumn{2}{c|}{PredCls}& \multicolumn{2}{c|}{SGCls}& \multicolumn{2}{c|}{SGDet}& \multicolumn{2}{c}{MEAN}\\ \cline{2-9} 
		\multicolumn{1}{c|}{}& \multicolumn{1}{c}{R@50/100} & \multicolumn{1}{c|}{mR@50/100} & \multicolumn{1}{c}{R@50/100} & \multicolumn{1}{c|}{mR@50/100} & \multicolumn{1}{c}{R@50/100} & \multicolumn{1}{c|}{mR@50/100} & \multicolumn{1}{c}{R-M} & \multicolumn{1}{c}{mR-M}\\ \hline
		
		{IMP}$^{\dag}$  \cite{suhail2021energy}  & 61.1 / 63.1 & 11.0 / 11.8 & 37.4 / 38.3 & 6.4 / 6.7 & 23.6 / 28.7 & 3.3 / 4.1 & 42.0 & 7.2 \\ 
		{GPS-Net}$^{\dag}$ \cite{li2021bipartite}  & 65.2 / 67.1 & 15.2 / 16.6 & 37.8 / 39.2 & 8.5 / 9.1 & 31.1 / 35.9 & 6.7 / 8.6 & 46.1 & 10.8 \\ 
		SG-CogTree \cite{yu2020cogtree} & 38.4 / 39.7 & 28.4 / 31.0 & 22.9 / 23.4 & 15.7 / 16.7 & 19.5 / 21.7 & 11.1 / 12.7 & 27.6 & 19.3\\ 
		BGNN \cite{li2021bipartite} & 59.2 / 61.3 & 30.4 / 32.9 & 37.4 / 38.5 & 14.3 / 16.5 & 31.0 / 35.8 & 10.7 / 12.6 & 43.9 & 19.6 \\ \hline
		VTransE \cite{tang2020unbiased} & \underline{\textbf{65.7}} / \underline{\textbf{67.6}} & 14.7 / 15.8 & \underline{38.6} / \underline{39.4} & 8.2 / 8.7 & \underline{29.7} / \underline{34.3} & 5.0 / 6.0 & \underline{45.9} & 9.7 \\
		VTransE + TDE \cite{tang2020unbiased} & 48.5 / 43.1 & 24.6 / 28.0 & 25.7 / 28.5 & 12.9 / 14.8 & 18.7 / 22.6 & 8.6 / 10.5 & 31.2 & 16.6 \\
		\textbf{VTransE + GCL} & 35.4 / 37.3 & \underline{34.2} / \underline{36.3} & 25.8 / 26.9 & \underline{20.5} / \underline{21.2} & 14.6 / 17.1 & \underline{13.6} / \underline{15.5} & 26.2 & \underline{23.5} \\\hline
		
		Motifs \cite{tang2020unbiased} &  \underline{65.2} / 67.0 & 14.8 / 16.1 & 38.9 / 39.8 & 8.3 / 8.8 & \underline{\textbf{32.8}} / \underline{\textbf{37.2}} & 6.8 / 7.9 & \underline{46.8} & 10.4 \\
		Motifs + Reweight \cite{chiou2021recovering} &  54.7 / 56.5 & 17.3 / 18.6 & 29.5 / 31.5 & 11.2 / 11.7 & 24.4 / 29.3 & 9.2 / 10.9 & 37.7 & 13.2 \\
		Motifs + TDE \cite{tang2020unbiased} & 46.2 / 51.4 & 25.5 / 29.1 & 27.7 / 29.9 & 13.1 / 14.9 & 16.9 / 20.3 & 8.2 / 9.8 & 32.1 & 16.8 \\
		Motifs + {PCPL}$^{\dag}$ \cite{chiou2021recovering}  & 54.7 / 56.5 & 24.3 / 26.1 & 35.3 / 36.1 & 12.0 / 12.7 & 27.8 / 31.7 & 10.7 / 12.6 & 40.4 & 16.4 \\
		Motifs + CogTree \cite{yu2020cogtree} & 35.6 / 36.8 & 26.4 / 29.0 & 21.6 / 22.2 & 14.9 / 16.1 & 20.0 / 22.1 & 10.4 / 11.8 & 26.4 & 18.1 \\
		Motifs + DLFE \cite{chiou2021recovering} & 52.5 / 54.2 & 26.9 / 28.8 & 32.3 / 33.1 & 15.2 / 15.9 & 25.4 / 29.4 & 11.7 / 13.8 & 37.8 & 18.7 \\
		Motifs + EMB \cite{suhail2021energy} & 65.2 / \underline{67.3} & 18.0 / 19.5 & \underline{39.2} / \underline{40.0} & 10.2 / 11.0 & 31.7 / 36.3 & 7.7 / 9.3 & 46.6 & 12.6 \\
		\textbf{Motifs + GCL} & 42.7 / 44.4 & \underline{36.1} / \underline{38.2} & 26.1 / 27.1 & \underline{20.8} / \underline{21.8} & 18.4 / 22.0 & \underline{16.8} / \underline{19.3} & 30.1 & \underline{25.5} \\\hline
		
		VCTree \cite{tang2020unbiased} & \underline{65.4} / \underline{67.2} & 16.7 / 18.2 & \underline{\textbf{46.7}} / \underline{\textbf{47.6}} & 11.8 / 12.5 & \underline{31.9} / \underline{36.2} & 7.4 / 8.7 & \underline{\textbf{49.2}} & 12.6 \\
		VCTree + Reweight \cite{chiou2021recovering} & 60.7 / 62.6 & 19.4 / 20.4 & 42.3 / 43.5 & 12.5 / 13.1 & 27.8 / 32.0 & 8.7 / 10.1 & 44.8 & 14.0 \\
		VCTree + TDE \cite{tang2020unbiased} & 47.2 / 51.6 & 25.4 / 28.7 & 25.4 / 27.9 & 12.2 / 14.0 & 19.4 / 23.2 & 9.3 / 11.1 & 32.5 & 16.8 \\
		VCTree + {PCPL}$^{\dag}$ \cite{chiou2021recovering}  & 56.9 / 58.7 & 22.8 / 24.5 & 40.6 / 41.7 & 15.2 / 16.1 & 26.6 / 30.3 & 10.8 / 12.6 & 42.5 & 17.0 \\
		VCTree + CogTree \cite{yu2020cogtree} &  44.0 / 45.4 & 27.6 / 29.7 & 30.9 / 31.7 & 18.8 / 19.9 & 18.2 / 20.4 & 10.4 / 12.1 & 31.8 & 19.8 \\ 
		VCTree + DLFE \cite{chiou2021recovering} & 51.8 / 53.5 & 25.3 / 27.1 & 33.5 / 34.6 & 18.9 / 20.0 & 22.7 / 26.3 & 11.8 / 13.8 & 37.1 & 19.5 \\
		VCTree + EMB \cite{suhail2021energy} & 64.0 / 65.8 & 18.2 / 19.7 & 44.7 / 45.8 & 12.5 / 13.5 & 31.4 / 35.9 & 7.7 / 9.1 & 47.9 & 13.5 \\ 
		\textbf{VCTree + GCL} & 40.7 / 42.7 & \underline{37.1} / \underline{39.1} & 27.7 / 28.7 & \underline{22.5} / \underline{23.5} & 17.4 / 20.7 & \underline{15.2} / \underline{17.5} & 29.6 & \underline{25.8} \\\hline
		
		SHA & 64.3 / 66.4 & 18.8 / 20.5 & 38.0 / 39.0 & 10.9 / 11.6 & 30.6 / 34.9 & 7.8 / 9.1 & 45.5 & 13.1 \\
		\textbf{SHA + GCL} & 35.1 / 37.2  & \textbf{41.6} / \textbf{44.1}  & 22.8 / 23.9 & \textbf{23.0} / \textbf{24.3} & 14.9 / 18.2 & \textbf{17.9} / \textbf{20.9} & 25.4 & \textbf{28.6} \\\hline

		\multicolumn{9}{c}{\multirow{2}{*}{Evaluation on GQA Dataset}}\\
		\multicolumn{9}{c}{}\\
		 
		\hline
		\multicolumn{1}{c|}{\multirow{2}{*}{Model}} & \multicolumn{2}{c|}{PredCls}& \multicolumn{2}{c|}{SGCls}& \multicolumn{2}{c|}{SGDet}& \multicolumn{2}{c}{MEAN}\\ \cline{2-9} 
		\multicolumn{1}{c|}{}& \multicolumn{1}{c}{R@50/100} & \multicolumn{1}{c|}{mR@50/100} & \multicolumn{1}{c}{R@50/100} & \multicolumn{1}{c|}{mR@50/100} & \multicolumn{1}{c}{R@50/100} & \multicolumn{1}{c|}{mR@50/100} & \multicolumn{1}{c}{R-M} & \multicolumn{1}{c}{mR-M}\\ \hline
		
		VTransE & 55.7 / 57.9 & 14.0 / 15.0 & 33.4 / 34.2 & 8.1 / 8.7 & 27.2 / 30.7 & 5.8 / 6.6 & 39.9 & 9.6 \\ 
		\textbf{VTransE + GCL} & 35.5 / 37.4 & 30.4 / 32.3 & 22.9 / 23.6 & 16.6 / 17.4 & 15.3 / 18.0 & 14.7 / 16.4 & 25.4 & 21.4 \\
		\hline
		
		Motifs & \textbf{65.3} / \textbf{66.8} & 16.4 / 17.1 & \textbf{34.2} / \textbf{34.9} & 8.2 / 8.6 & \textbf{28.9} / \textbf{33.1} & 6.4 / 7.7 & \textbf{43.9} & 10.9 \\ 
		\textbf{Motifs + GCL} & 44.5 / 46.2 & 36.7 / 38.1 & 23.2 / 24.0 & 17.3 / 18.1 & 18.5 / 21.8 & 16.8 / 18.8 & 29.7 & 24.2 \\
		\hline
		
		VCTree & 63.8 / 65.7 & 16.6 / 17.4 & 34.1 / 34.8 & 7.9 / 8.3 & 28.3 / 31.9 & 6.5 / 7.4 & 43.1 & 10.5 \\
		\textbf{VCTree + GCL} & 44.8 / 46.6 & 35.4 / 36.7 & 23.7 / 24.5 & 17.3 / 18.0 & 17.6 / 20.7 & 15.6 / 17.8 & 29.6 & 23.6 \\
		 \hline
		 
		SHA & 63.3 / 65.2 & 19.5 / 21.1 & 32.7 / 33.6 & 8.5 / 9.0 & 25.5 / 29.1 & 6.6 / 7.8 & 41.6 & 12.1 \\
		\textbf{SHA + GCL} & 42.7 / 44.5 & \textbf{41.0} / \textbf{42.7} & 21.4 / 22.2 & \textbf{20.6} / \textbf{21.3} & 14.8 / 17.9 & \textbf{17.8} / \textbf{20.1} & 27.3 & \textbf{27.3} \\
		 \hline

	\end{tabular}
	\caption{Detailed performance comparison of different methods on PredCls, SGCls, and SGDet tasks of both VG150 and GQA200 with respect to R@50/100 (\%), mR@50/100 (\%), and their mean (\%). R-M and mR-M denote the mean on all three tasks over R@50/100 and mR@50/100, respectively. The optimal results from the same baseline (\ie, VTransE, Motifs and VCTree) in VG150 are underlined. The global optimal results over all the methods in VG150 and GQA200 are in bold. The superscript $\dag$ denotes that the method is reproduced. Note that all the methods are implemented on the same object detector, \ie, a pre-trained Faster R-CNN with ResNeXt-101-FPN.}
	\label{result_all}
\end{table*}

\end{document}